%% file: main.tex
\documentclass[10pt,twocolumn,letterpaper]{article}

\usepackage[pagenumbers]{cvpr}

\usepackage{multirow}
\usepackage{tabularx}

\usepackage{graphicx}
\usepackage[table]{xcolor}
\usepackage{hhline}
\usepackage{animate}
\usepackage{ulem}


\definecolor{cvprblue}{rgb}{0.21,0.49,0.74}
\usepackage[pagebackref,breaklinks,colorlinks,citecolor=cvprblue]{hyperref}
\usepackage{booktabs}
\def\paperID{18} 
\def\confName{CVPR}
\def\confYear{2024}

\title{NeRFFaceSpeech: One-shot Audio-driven 3D Talking Head Synthesis via Generative Prior}

\author{%
Gihoon Kim$^{1}$\qquad Kwanggyoon Seo$^{2}$\qquad Sihun Cha$^{2}$\qquad Junyong Noh$^{2}$\\
\vspace{-1mm}
{\normalsize $^{1}$Seoul National University \quad $^{2}$KAIST} \\
{\tt\small gihoon.kim@snu.ac.kr\qquad \{skg1023, chacorp, junyongnoh\}@kaist.ac.kr}
\\{\normalsize \href{https://rlgnswk.github.io/NeRFFaceSpeech_ProjectPage/}{Project Page}}
\vspace{-3mm}
}

\begin{document}
\twocolumn[{%
\renewcommand\twocolumn[1][]{#1}%
\maketitle
\begin{center}
    \vspace{-5mm}
    \centering
    \captionsetup{type=figure}
    \includegraphics[width=0.88\textwidth]{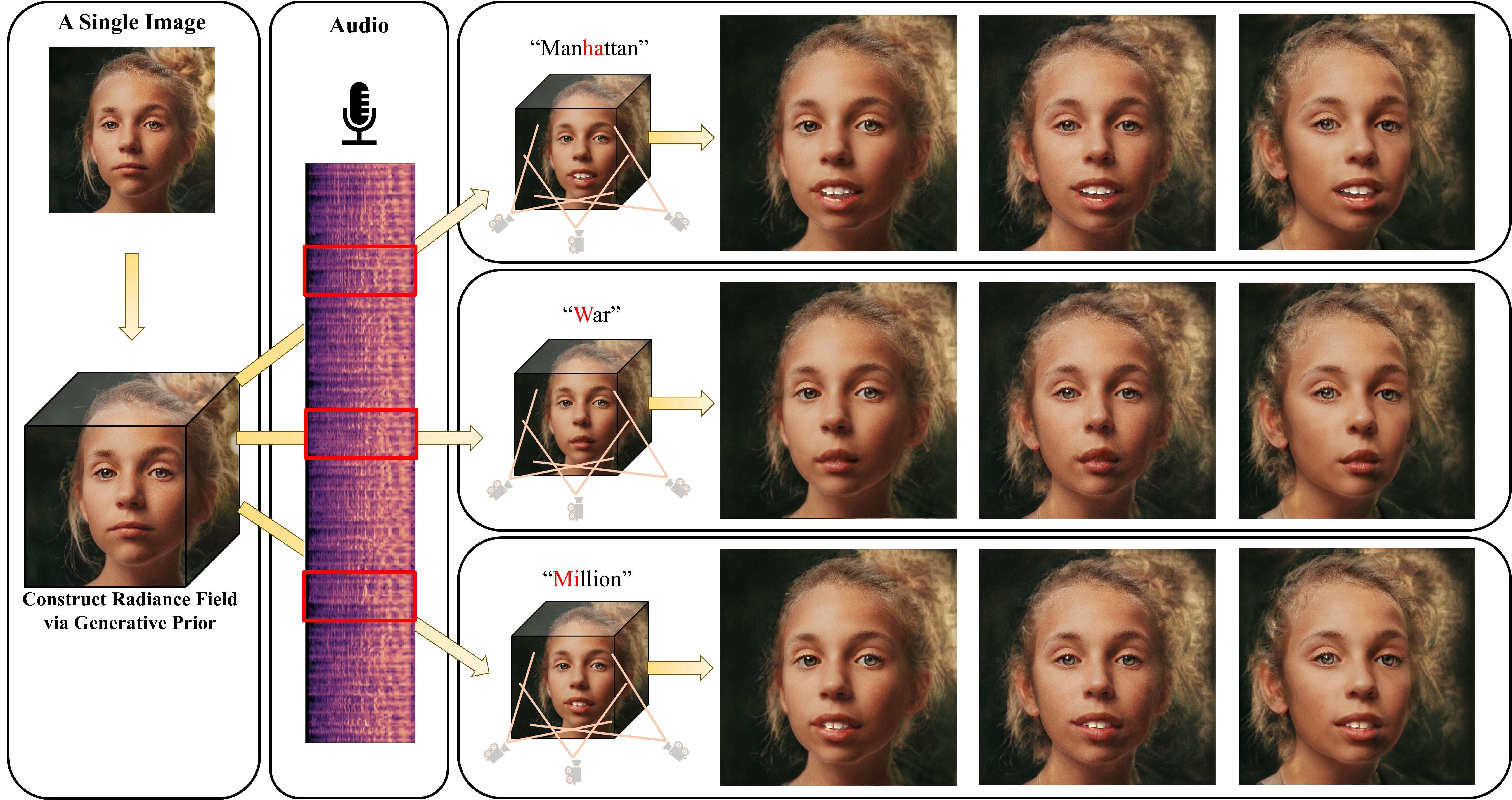}
    \caption{Our approach, NeRFFaceSpeech, constructs a 3D-aware facial feature space from a single image utilizing generative priors and incorporates audio-driven dynamics via ray deformation, enabling the synthesis of talking heads from novel viewpoints.}
    \label{fig:teaser}
\end{center}%
}]

\input{sec/0_abstract}    
\input{sec/1_intro}
\input{sec/2_related}
\input{sec/3_preliminary}
\input{sec/4_method}

\input{sec/5_experiment}
\input{sec/6_conclusion}
\clearpage
{
    \small
    \bibliographystyle{ieeenat_fullname}
    \nocite{*}
    \bibliography{main}
}



\end{document}

%% file: sec/0_abstract.tex
\begin{abstract}
\vspace{-3mm}
Audio-driven talking head generation is advancing from 2D to 3D content. Notably, Neural Radiance Field (NeRF) is in the spotlight as a means to synthesize high-quality 3D talking head outputs. Unfortunately, this NeRF-based approach typically requires a large number of paired audio-visual data for each identity, thereby limiting the scalability of the method. Although there have been attempts to generate audio-driven 3D talking head animations with a single image, the results are often unsatisfactory due to insufficient information on obscured regions in the image. In this paper, we mainly focus on addressing the overlooked aspect of 3D consistency  in the one-shot, audio-driven domain, where facial animations are synthesized primarily in front-facing perspectives. We propose a novel method, NeRFFaceSpeech, which enables to produce high-quality 3D-aware talking head. Using prior knowledge of generative models combined with NeRF, our method can craft a 3D-consistent facial feature space corresponding to a single image. Our spatial synchronization method employs audio-correlated vertex dynamics of a parametric face model to transform static image features into dynamic visuals through ray deformation, ensuring realistic 3D facial motion. Moreover, we introduce LipaintNet that can replenish the lacking information in the inner-mouth area, which can not be obtained from a given single image. The network is trained in a self-supervised manner by utilizing the generative capabilities without additional data. The comprehensive experiments demonstrate the superiority of our method in generating audio-driven talking heads from a single image with enhanced 3D consistency compared to previous approaches. In addition, we introduce a quantitative way of measuring the robustness of a model against pose changes for the first time, which has been possible only qualitatively.
\end{abstract}

%% file: sec/1_intro.tex
\section{Introduction}
\vspace{-1mm}
Generating a talking head animation from audio data has emerged as a pivotal technology, gaining substantial attention in applications such as digital human and visual dubbing within the film and gaming industries. Furthermore, the landscape of content consumption has been radically transformed by the advent of next-generation media platforms, such as Virtual Reality (VR) and Augmented Reality (AR). Given these changes, the ability to create convincing 3D facial animations synchronized with audio is not merely a technological advancement; it is an essential component in driving user engagement that is central to the success of contemporary digital experiences.

In this context, Neural Radiance Field (NeRF) \cite{mildenhall2020nerf} has garnered attention due to its capability to successfully generate 3D content. In the field of audio-driven facial animation, there has been an active progression of research integrating audio features within NeRF \cite{guo2021adnerf, liu2022semantic, yao2022dfa, shen2022learning, li2023efficient, tang2022radnerf}. Such studies have demonstrated that a network with neural rendering can be trained in conjunction with audio information, yielding 3D-aware results. However, a notable limitation of these frameworks is their requirement for extensive audio-image pairs, specifically speech videos, compounded by the need for suitable multi-view images. The need to acquire datasets of this type and size poses a considerable constraint on the scalability of such frameworks.

Conversely, there are studies \cite{suwajanakorn2017synthesizing, prajwal2020lip, kr2019towards, zhang2023sadtalker, zhou2020makelttalk, wang2021audio2head, wang2022one} that operate on a single input image. These methods have attempted to train models to capture generic audio-visual correlations. This advancement has demonstrated the capability to produce convincing talking head animations from a single image in inference time. Nevertheless, due to the inherent constraints of deriving multifaceted information from a single image, these methods still reveal deficiencies when attempting to generate a broad spectrum of poses.

To address 3D information hurdles with a single image, we turn our attention to harnessing the capacity of generative models. Recent advancements \cite{chan2021pi, schwarz2020graf, niemeyer2021giraffe, gu2022stylenerf, Chan2022eg3d, deng2022gram, xiang2022gramhd} in facial image synthesis have demonstrated the feasibility of combining the established structure of 2D image generation models like StyleGAN \cite{karras2019style, karras2020analyzing, Karras2021} and neural rendering pipeline to generate 3D-aware images. Taking advantage of the feature space embedded within these generative models, we can construct a comprehensive 3D head feature space, even when constrained to a single-viewpoint image as input through GAN inversion \cite{roich2022pivotal}. Moreover, recent research \cite{pan2023drag, shi2023dragdiffusion} highlights the potential of explicitly handling points within the feature space for interactive editing.

In this paper, we introduce NeRFFaceSpeech, a novel framework capable of synthesizing 3D-aware talking head corresponding to audio and a single image input, using generative prior of the backbone model as shown in Figure \ref{fig:teaser}. Following this, we design a method to map the vertices of a parametric face model to the 3D feature space constructed from a single image, enabling appropriate facial movement through ray deformation reflected in audio-corresponding vertices. This approach ensures consistency in the conversion of pose changes, which were implicitly handled in previous methods, by explicitly transforming the features.

Utilizing a single image as input often leads to insufficient information for facial animation, especially in the mouth region. To address this, we introduce an inpainting network, named LipaintNet, which is trained in a self-supervised manner to add missing information inside the mouth. Because LipaintNet does not require  additional data or retraining of the entire backbone model \cite{sun2023next3d}, it can avoid unintended alteration that is caused by catastrophic forgetting, which may lead to degraded quality of the generated results.

We assess the robustness of audio-driven talking heads to pose changes by measuring the differences in outcomes using the horizontally flipped input images. This approach offers a novel way to measure the robustness in pose changes quantitatively, which has been possible only by a qualitative assessment. Through extensive experiments, we demonstrate the strength of our method in multi-view audio-driven talking head generation, even when receiving only a single image by harnessing the prior knowledge of generative models.

%% file: sec/2_related.tex
\section{Related Work}
\vspace{-1mm}
\subsection{Audio-driven Talking Head Generation}
\vspace{-1mm}
Audio-driven video synthesis, particularly focusing on talking head animations, is a challenging task due to the necessity of mastering the correlations of different modality. Early studies \cite{suwajanakorn2017synthesizing, prajwal2020lip, kr2019towards} were primarily directed towards audio-lip synchronization. Among these, Wav2Lip \cite{prajwal2020lip} and LipGAN \cite{kr2019towards} learned the multi-modality between audio and lip through the use of the GAN paradigm \cite{goodfellow2014generative} and contrastive learning. However, the output from these work often appears unnatural, presenting static images with moving lips.

In response, following studies expanded their scope to model comprehensive facial movements. By learning the correspondence between 
the audio and visual information using landmarks \cite{zhou2020makelttalk}, a warping map \cite{wang2021audio2head, wang2022one, zhou2021pose, yin2022styleheat} or parametric models \cite{ren2021pirenderer, zhang2023sadtalker}, these approaches manage to implicitly link audio-visual correspondence using neural networks. Particularly, SadTalker \cite{zhang2023sadtalker} attained state-of-the-art performance by employing Wav2Lip to learn motion parameters from audio and bridging its relationship with a 3D-aware renderer \cite{wang2021one}. Nonetheless, a single image fails to capture sufficient perspectives of an individual. Consequently, these methods produce unstable outcomes, which becomes especially evident when attempting to generate the arbitrary view of an input image.

In light of these difficulties, constructing a implicit 3D space through NeRF \cite{mildenhall2020nerf} has been explored as a feasible solution. These methods showcase the capability to learn dynamic scenes corresponding to a given audio feature \cite{guo2021adnerf} with supplementary feature \cite{yao2022dfa, liu2022semantic, ye2023geneface} such as semantic information or structural efficient representation \cite{tang2022radnerf, li2023efficient}. However, despite efforts to reduce the required data \cite{shen2022learning}, the drawback of requiring numerous audio-image pairs for each identity remains unresolved. 

\begin{figure*}[t]
  \centering
  \vspace{-3mm}
   \includegraphics[width=0.95\linewidth]{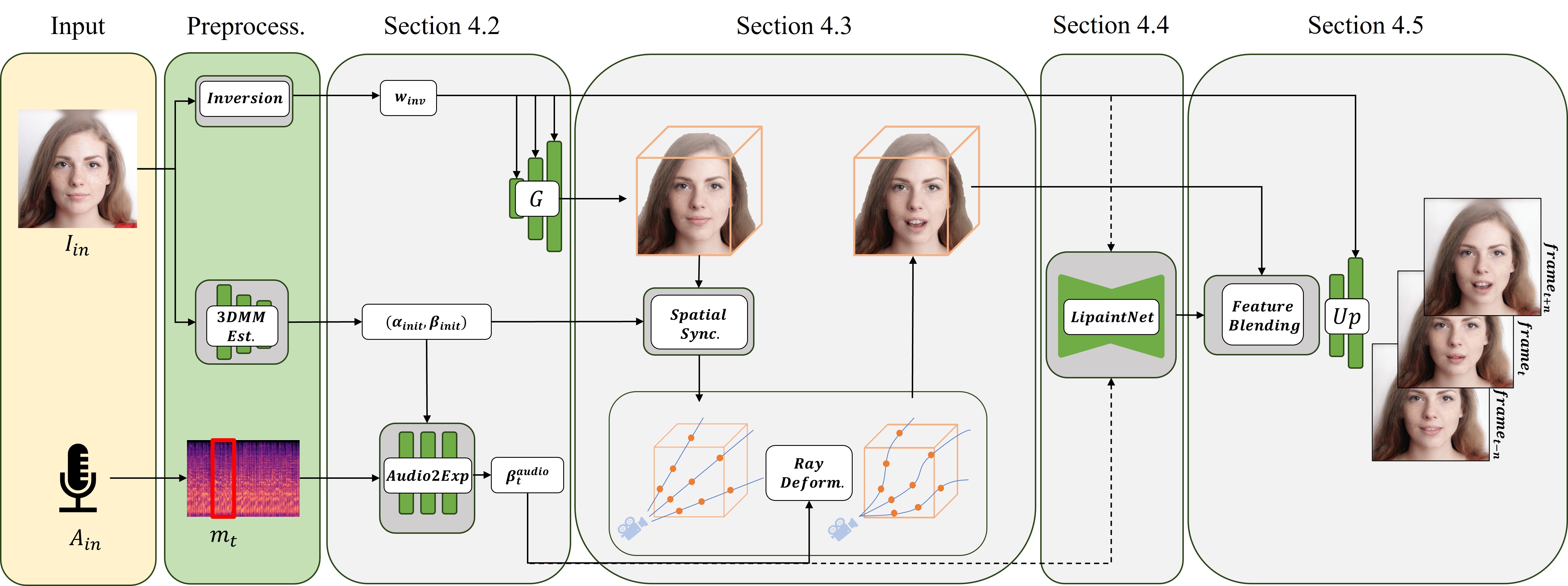}
   \caption{Overall pipeline of our method: Given a single image and an audio input, the preprocessing stage extracts 3DMM parameters and 3D features in the NeRF space. Subsequent spatial synchronization between 3DMM and feature space enables the reflection of expression vertex changes within the feature space. These changes are computed in accordance with the audio time step $t$, resulting in the generation of facial movements. The final output frame emerges from feature blending, where the deformed features are merged with the inner-mouth details generated by the proposed LipaintNet.}
   \vspace{-3mm}
   \label{fig:Overview}
\end{figure*}

\subsection{Dynamic Neural Scene Representation}
\vspace{-1mm}
Recent advancements in Neural Scene Representation techniques not only focus on effective representation of static scenes \cite{mildenhall2020nerf, zhang2020nerf++, barron2021mip, yariv2021volume, muller2022instant} but also increasingly explore capturing active elements within dynamic scenes. These approaches aim to identify suitable representations for dynamic scenes \cite{li2022neural, park2021nerfies, park2021hypernerf, Lombardi21, Lombardi2019}. Furthermore, by embedding facial features such as landmarks and parameters of a facial model \cite{Gao2022nerfblendshape, cao2022authentic, grassal2022neural, hong2022headnerf, wang2021learning, gafni2021dynamic, zeng2022fnevr, peng2021animatable} or audio features \cite{guo2021adnerf, yao2022dfa, liu2022semantic, ye2023geneface, tang2022radnerf, li2023efficient, shen2022learning}, these methods facilitate the representation of dynamics in the desired direction through the learned features. In addition, by identifying embedding spaces corresponding to specific conditions, another methods \cite{ma2023otavatar, Xu2023LatentAvatar, tang2022explicitly} exhibit a way to represent the motion. 

These implicit learning strategies may lead to attributes that can become unintentionally entangled, causing undesired concurrent changes in specific details \cite{sun2022controllable}. Other studies proposed an approach to explicitly deforming the scene to intuitively represent dynamics \cite{jambon2023nerfshop, pumarola2021d, yuan2022nerf}. These methods leverage the point-wise projection mechanism of neural rendering pipeline for deriving dynamics from existing static scenes by transforming to each radiance. Moreover, by associating keypoints \cite{chen2023implicit, tretschk2021non} or mesh vertices \cite{athar2022rignerf, zheng2022imface, sun2022controllable, xu2023omniavatar} to its radiance fields, such methods enable the articulation of desired dynamics within static scenes and provide interactive representations. Building on this approach, our work also capitalizes on these merits to effectively utilize audio-driven dynamics within radiance fields.

%% file: sec/3_preliminary.tex
\section{Preliminary}
\vspace{-1mm}
\subsection{Style-Based Neural Rendering}
\vspace{-1mm}
Building upon the Neural Radiance Field (NeRF) \cite{mildenhall2020nerf}, our backbone, StyleNeRF \cite{gu2022stylenerf}, integrates a generative paradigm \cite{goodfellow2014generative}. In this approach, a latent vector $z$ is sampled from a gaussian distribution $\mathcal{Z}$. Then, a mapping network $f(\cdot)$ produces a style vector $w$ from $z$, $w = f(z)$. Our backbone fundamentally follows the NeRF pipeline, formulating the volumetric scene as a continuous function. It maps 3D coordinate $p=(x, y, z)$ with  positional encoding $\zeta$ \cite{tancik2020fourier} to a volume density $\sigma$ and radiance $c$. For a given location $p$ and viewing direction $d$: 
\begin{equation}
\label{GenerativeMLP}
\mathcal{F}(\zeta(p),\zeta(d), w) \rightarrow (c, \sigma),
\end{equation}
where $\mathcal{F}$ is a multi-layer perceptron. The style vector $w$ is integrated into the rendering pipeline, serving as an additional input to the neural network that parameterizes the volume. In the rendering process, along a camera ray $r(t)=o+t\cdot d$ (where $o$  represent the origin of the ray and $d$ represents the viewing direction), the continuous representation is integrated:
\begin{equation}
\label{eqn:marching}
C(r) = \int_0^{\infty} \sigma(r(t)) c(r(t), d) T(r(t)) dt,
\end{equation}
where $T(t) = \exp\left(-\int_0^t \sigma(r(s)) ds\right)$ represents the accumulated transmittance up to sampled points $t$. Following the adapted framework $\mathcal{F}$, we can obtain a  feature map $\phi_{w}$ via volume rendering based on Eq.~\ref{eqn:marching}. Subsequently, by leveraging an upsampler grounded in the StyleGAN architecture \cite{karras2020analyzing}, this process can generate a high-resolution image as $I^{Up}_{w}=Upsampler(\phi_{w}, w)$.

\subsection{Parametric Face Model}
\vspace{-1mm}
The 3D Morphable Model (3DMM) \cite{Blanz99amorpahalbe} is a parametric representation of the human facial structure in three-dimensional space. Originating from a set of 3D facial scans, the model encodes both the shape and appearance variations. In this work, we focus specifically on the shape information, from which vertex positions can be extracted. The shape is represented as:
\begin{equation}
\label{eqn:3dmm}
S=\overline{S}+\alpha S_{id}+\beta S_{exp}.
\end{equation}
In this representation, $\overline{S}$ is the mean value of the shape, while $S_{id}$ and $S_{exp}$ are the basis for identity and expression, which are computed using Principal Component Analysis (PCA). The coefficients $\alpha$ and $\beta$ function as weights for these basis components. 

%% file: sec/4_method.tex
\section{Method}
\vspace{-1mm}
In this section, we introduce the proposed pipeline as shown in Figure \ref{fig:Overview}. First, we perform a preprocessing step operating on the input image for initial setting of the pipeline (\cref{subsec:Preprocessing}). Next, we obtain the expression parameters corresponding to the audio input (\cref{subsec:Audio-to-3DMM}). We then leverage ray deformation through displacements informed by audio-driven shape variations, facilitating the creation of talking head animations building on the aligned 3DMM vertices with NeRF features (\cref{subsec:Rendering}). In addition, recognizing the limitations of ray deformation, especially in capturing the inner-mouth area, we utilize a novel self-supervised inpainting network (\cref{subsec:LipaintNet}). In the final step, all information extracted from the feature space is blended to produce an appropriate result (\cref{subsec:Blending}).

\subsection{Preprocessing}\label{subsec:Preprocessing}
\vspace{-1mm}
Given an input image $I_{in}$, we employ a 3DMM estimator \cite{deng2019accurate} to predict its initial parameters, $\alpha^{init}$ and $\beta^{init}$ for $S^{init}$ in Eq.~\ref{eqn:3dmm} as well as the input pose. Recognizing the significance of accuracy, we further refine these parameters by optimizing them using gradient descent, ensuring a more precise matching \cite{ascust2023}. We then employ Pivotal Tuning Inversion (PTI) \cite{roich2022pivotal} to obtain the inverted style vector $w_{inv}$ which undergoes projection and tuning for 1000 steps each. This inversion process enables to adapt $I_{in}$ to the latent space of our backbone \cite{gu2022stylenerf}. We transform the audio input $A_{in}$ into a mel spectrogram $m$ with dimensions $T_a \times D$, where $T_a$ denotes the audio time and $D$ represents the mel spectrogram dimension as specified in Wav2Lip \cite{prajwal2020lip} to ensure a standard and consistent representation for a downstream task (\cref{subsec:Audio-to-3DMM}).

\subsection{Audio to Expression}\label{subsec:Audio-to-3DMM}
\vspace{-1mm}
The Audio2Exp module, introduced in SadTalker \cite{zhang2023sadtalker}, is trained to learn the correlation between audio data and the 3DMM expression weights $\beta$ in Eq.~\ref{eqn:3dmm}. This model, leveraging the pretrained Lip-Sync network proposed in Wav2Lip, is designed to produce appropriate expression coefficients for lip motion corresponding to the audio. The architecture accepts a segment $m_{t}$ of a mel-spectrogram for a specific duration $t\in T_a$. Along with $\beta^{init}$ as a basis expression input, Audio2Exp module then outputs the audio-related expression parameters $\beta^{audio}$. The model can be represented as:

\begin{equation}
\label{eqn:audio2lip}
\beta^{audio}_{t} = Audio2Exp(m_{t}, \beta).
\end{equation}

For a given audio input data, we can determine the corresponding coefficient $\beta^{audio}_{t}$ at each time step $t$ through Eq.~\ref{eqn:audio2lip}. With these coefficients, based on Eq.~\ref{eqn:3dmm}, the facial shape $S^{audio}$ corresponding to the audio information can be constructed as follows:
\begin{equation}
\label{eqn:audio3dmm}
S^{audio}_{t}=\overline{S}+\alpha^{init} S_{id}+\beta^{audio}_{t} S_{exp}.
\end{equation}

\subsection{Spatial Synchronization via Ray Deformation}\label{subsec:Rendering}
\vspace{-1mm}
This section is about a synchronization of the spatial representation between the audio-driven 3DMM vertices and the canonical space of facial features. Specifically, vertices $V^{init}$ extracted from $S^{init}$ are coherently aligned and projected onto the 3D feature space represented by Eq.~\ref{GenerativeMLP} with $w_{inv}$. To achieve a precise mapping, we employ a Nearest Neighbor matching. The alignment process is conducted on the XY-plane corresponding to a frontal-view perspective to ensure no vertices are overlapped or mapped twice. Therefore, associating each vertex $(x_{i},y_{i}) \in V^{init}$ with its closest counterpart points $(x_{n},y_{n})$ sampled $\phi_{w_{inv}}$, are calculated by
\begin{equation}
\label{eqn:NN}
NN(x_i, y_i) = argmin_{(x^{'}_{i},y^{'}_{i})} \| (x^{'}_{i},y^{'}_{i}) - (x_{n},y_{n}) \|,
\end{equation}
where $(x^{'}_{i},y^{'}_{i})$ represents the coordinates of the vertex transformed to align with the scale of the neural radiance coordinate system.

This approach guarantees that the spatial relationships from the 3DMM are preserved and translated accurately to the feature space of our backbone, with each vertex matched to its most proximate feature. Thus, it implies that the two distinct features, 3DMM and the feature space, are interconnected, ensuring mutual reflection of variations between them. Consequently, we can now freely take ray deformation based on 3DMM vertex movements corresponding to audio information by computing a displacement $\Delta V$ between the vertices as follows: $\Delta V_{t}=V^{init}-V^{audio}_{t}$, where vertices $V^{audio}_{t}$ are calculated using Eq.~\ref{eqn:audio3dmm}.

Conceptually similar to how facial muscles move to create expressions, movement in response to audio in the static feature space can produce animation. This can be represented as follows:
\begin{equation}
\label{eqn:raydeform}
\mathcal{F}(\zeta(p+\Delta p),\zeta(d), w_{inv}) \rightarrow c, \sigma,
\end{equation}
where $\Delta p \in \Delta V$. Consequently, each radiance
$r(t)$ evolves to $r^{*}(t)$ due to the shift by $\Delta p$, leading to a modified volume rendering described by: $C(r^{*}) = \int_0^{\infty} \sigma(r^{*}(t)) c(r^{*}(t), d) T(r^{*}(t)) dt$ from Eq.~\ref{eqn:marching}.

In practice, we apply a momentum to the vertex displacement to promote more natural movements, instead of simply reflecting the displacement at time step $t$. By incorporating a constant $\alpha_{m}$, the input $\beta$ in Eq.~\ref{eqn:audio2lip} has the contribution for displacement momentum from the previous time step, which is formulated as $\beta = (1-\alpha_m)\cdot\beta^{init} + \alpha_m\cdot\beta^{audio}_{t-1}$. To further enhance the dynamism of the scene, we adjust a constant $\lambda_{exp}$  to modulate the scale of facial movements. This adjustment is applied by multiplying $\lambda_{exp}$ with $\beta^{audio}_{t}$ from Eq.~\ref{eqn:audio3dmm} and results in more fluid and emphasized movements within the rendered output as detailed in Table \ref{tab:ablation_table}. With this modification, the final displacement $\Delta V$ is computed, leading to the formation of the deformed feature $\phi^{*}_{D}$, shaped by the audio input.

\subsection{LipaintNet}\label{subsec:LipaintNet}
\vspace{-1mm}
While the established correspondence allows for the representation of appropriate facial movements via ray deformation, this approach inherently needs to generate additional information. This demand becomes particularly evident when the mouth opens, revealing the inside. To address this, we present LipaintNet, designed explicitly to supplement these missing details. Directly searching for desired latents is challenging, as adjusting in the latent space often inadvertently modifies unintended details \cite{sun2022controllable}. We avoid this by pivoting to a more tractable task solely focused on capturing internal mouth details. The objective of LipaintNet is to convert an input identity latent $w_{id}$ into an expression-compatible latent $w_{exp}$ in response to a target expression parameter $\beta_{target}$. This can be expressed as follows:
\begin{equation}
\label{eqn:inpaint}
w_{exp}=LipaintNet(\beta_{target},w_{id}).
\end{equation}
The output of this process is fed into the backbone model as an input, creating inner-mouth features required in \cref{subsec:Blending}.

We trained this model in a self-supervised manner, leveraging the generative prior of the backbone capacity. Given that sampling from the latent space $\mathcal{Z}$ allows for virtually "infinite" image generation, we exploit this advantage to train our LipaintNet without the need for additional data. First, a 3DMM estimator \cite{deng2019accurate} is utilized to extract facial coefficients and landmarks from randomly sampled images. Next, we derive pseudo Ground Truth (GT) coefficients $\alpha_{target}, \beta_{target} \in \mathcal{S}_{target}$ and landmarks $ldm_{target}$ by applying random augmentation. As in Eq.~\ref{eqn:inpaint}, LipaintNet takes the $\beta_{target}$ and a sampled $w_{id}$, ensuring that the facial coefficients $\widehat{\mathcal{S}}_{out}$ and landmarks $\widehat{ldm}_{out}$ extracted from the final generated image from $w_{exp}$ align with the pseudo GT values.

The objective function for training LipaintNet comprises two key losses: the landmark loss $\mathcal{L}_{ldm}$ and  the coefficient loss $\mathcal{L}_{3DMM}$. These are calculated using the L2 distance metric represented as follows:
\begin{equation}
\label{eqn:GAN}
\mathcal{L}_{GAN} = -\log [\mathcal{D}(\mathcal{G}(\widehat{w}_{exp}))]
\end{equation}
and
\begin{equation}
\label{eqn:ldm}
\mathcal{L}_{ldm} = {||ldm_{target}-\widehat{ldm}_{out}||}_{2},
\end{equation}
respectively. To further enhance the perceptual quality, we leverage a pretrained discriminator $\mathcal{D}$ \cite{gu2022stylenerf}. Thus, the GAN loss, $\mathcal{L}_{GAN}$, is also utilized to train the model using Binary Cross Entropy as shown:

\begin{equation}
\label{eqn:exp}
\mathcal{L}_{3DMM} = {||\mathcal{S}_{target}-\widehat{\mathcal{S}}_{out}||}_{2}.
\end{equation}
The complete loss function for LipaintNet can be written as follows: $\mathcal{L}_{total}=\lambda_{ldm}\cdot\mathcal{L}_{ldm}+\lambda_{exp}\cdot\mathcal{L}_{exp}+\lambda_{GAN}\cdot\mathcal{L}_{GAN}$. We employ expression parameters for the loss calculation as well as identity parameters as a regularizer, ensuring the preservation of the identity. Further details about the training are elaborated in Sec. \textcolor{red}{E} in the supplementary material.

In the inference phase, the trained LipaintNet takes as input the $w_{inv}$ derived from the input image and the parameter $\beta^{audio}_{t}$. It outputs $w^{audio}_{exp}$, which is close to $w_{inv}$ and corresponds to the audio expression $\beta^{audio}$. This enables us to utilize appropriate mouth feature in a feed-forward manner at each time step $t$ as follows: $w^{audio_t}_{exp}=LipaintNet(\beta^{audio}_{t},w_{inv})$.

\subsection{Feature Blending}\label{subsec:Blending}
\vspace{-1mm}
In the final step, we blend the deformed feature $\phi^{*}_{D}$ representing facial movements (\cref{subsec:Rendering}) with the feature $\phi^{audio}_{exp}$ obtained from $w^{audio}_{exp}$ (\cref{subsec:LipaintNet}). Both features are converted to 2D features based on camera parameters $\psi$, as illustrated in Eq.~\ref{eqn:marching}. We utilize the mask $\mathcal{M}$, which represents the mouth area as determined by the pose $\psi$ derived from the 3DMM parameters, to calculate the blended feature as:
\begin{equation}
\label{eqn:inpaint}
\phi^{*}_{bld}\mid_{\psi}=\phi^{*}_{D}\mid_{\psi}\odot(1-\mathcal{M}\mid_{\psi}) +(\phi^{audio}_{exp}\mid_{\psi}\odot\mathcal{M}\mid_{\psi}).
\end{equation}
In practice, to alleviate the issue of generating masks that do not transition smoothly from frame to frame at each time step $t$, we utilize the average of the previous $n$-frames for the mask $\mathcal{M}_t$, computed as
\begin{equation}
\label{eqn:avgm}
\mathcal{M}_t = \frac{1}{n} \sum_{i=1}^{n} \mathcal{M}_{t-i}.
\end{equation}
Consequently, through the aforementioned process, we derive $\phi^{*}_{bld}\mid_{\psi}$ and employ the upsampler from the backbone model to produce the final image corresponding to the audio, expressed as:
\begin{equation}
\label{eqn:final}
I_{Final}=Upsampler(\phi^{*}_{bld}\mid_{\psi}, w_{inv}).
\end{equation}

%% file: sec/5_experiment.tex
\section{Experiment}
\vspace{-1mm}
\subsection{Implementation Details}
\vspace{-1mm}
We ensured deterministic sampling points for neural rendering to the reduced artifacts stemming from frame-to-frame discrepancies. We set $\alpha_m$ as 0.5 and $\lambda_{exp}$ as 1.5 and $n=7$ in Eq.\ref{eqn:avgm}. The proposed LipaintNet consists of fully connected layers, which adds residuals to the given latent input $w$. For the weights of the objective functions during LipaintNet training, we set $\lambda_{ldm}=1.0$, $\lambda_{3DMM}=1.0$, and $\lambda_{GAN}=0.1$. The augmentation for the expression parameter described in \cref{subsec:LipaintNet} was performed with uniform sampling within the [-1.0, 1.0] range. Our LipaintNet model underwent 300k iterations of training using the Adam optimizer \cite{kingma2014adam} with a learning rate of 1e-5, with a batch size of 8 on an NVIDIA RTX A5000 GPU.

\subsection{Dataset}
\vspace{-1mm}
For our experiments, we utilized the HDTF Dataset \cite{zhang2021flow} for both images and audio, and the Unplash Dataset \cite{zhu2022hairnet} for 120 high-resolution images sourced online. Following the settings established by Sadtalker \cite{zhang2023sadtalker}, we used 400 videos from the HDTF Dataset, whose duration is 8 seconds each. For each video, the initial frame was used as input along with the 8-second audio segment. These images served as our visual input, while the HDTF Dataset provided the accompanying audio input. To cater to the requirements of the Audio2Exp module from Wav2Lip \cite{prajwal2020lip}, we adjusted the audio's sampling rate of the audio to 16kHz and transformed it into a mel-spectogram. 

\begin{table}[t]
    \centering
    \resizebox{0.45\textwidth}{!}{
    \begin{tabular}{lccccc|ccccc}
        \toprule
        \multirow{2}{*}{Method} & \multicolumn{5}{c|}{Unplash} & \multicolumn{5}{c}{HDTF} \\
        \cline{2-6}\cline{7-11}
         & FID$\downarrow$ & CSIM$\uparrow$ & CPBD$\uparrow$ & LSE-D$\downarrow$ & LSE-C$\uparrow$ & FID$\downarrow$ & CSIM$\uparrow$ & CPBD$\uparrow$ & LSE-D$\downarrow$ & LSE-C$\uparrow$
         \\
        \hline
        \rowcolor{gray!20}
        Ground Truth & 0.000 & 1.000 & 0.639 & N/A & N/A & 0.000 & 1.000 & 0.301 & 6.843 & 8.398 \\
        Wav2Lip \cite{prajwal2020lip} & \textbf{48.164} & \textbf{0.893} & 0.472 & 7.079 & \textbf{8.400} &  \textbf{13.541} & 0.933 & \textbf{0.314} & \textbf{6.329} & \textbf{9.148} \\
        MakeItTalk \cite{zhou2020makelttalk} & 52.585 & 0.833 & 0.134 & 9.901 & 5.063 & 20.591 & 0.881 & 0.274 & 9.889 & 5.139 \\
        PC-AVS \cite{zhou2021pose} & 50.223 & 0.620 & 0.157 & \textbf{7.018} & 8.007 & 39.764 & 0.699 & 0.148 & 6.501 & 8.738 \\
        Audio2Head \cite{wang2021audio2head} & 56.965 & 0.581 & 0.316 & 7.529 & 7.374 & 20.060 &  0.752 & 0.272 & 7.529 & 7.374 \\
        SadTalker \cite{zhang2023sadtalker} & 56.978 & 0.886 & 0.184 & 7.989 & 6.913 &  15.597 & \textbf{0.937} & 0.164 & 7.618 & 7.463 \\
        Ours & 50.463 & 0.747 & \textbf{0.581} & 8.869 & 5.941 & 36.994 & 0.826 & 0.225 & 9.194 & 5.729 \\
        \toprule
    \end{tabular}
    }
    \vspace{-3mm}
    \caption{Quantitative comparison of the general performance using Unplash and HDTF Datasets (Bold: highest).}
    \vspace{-3mm}
    \label{tab:comparison_Unpalsh}
\end{table}

\begin{figure}[t]
  \centering
   \includegraphics[width=0.40\textwidth]{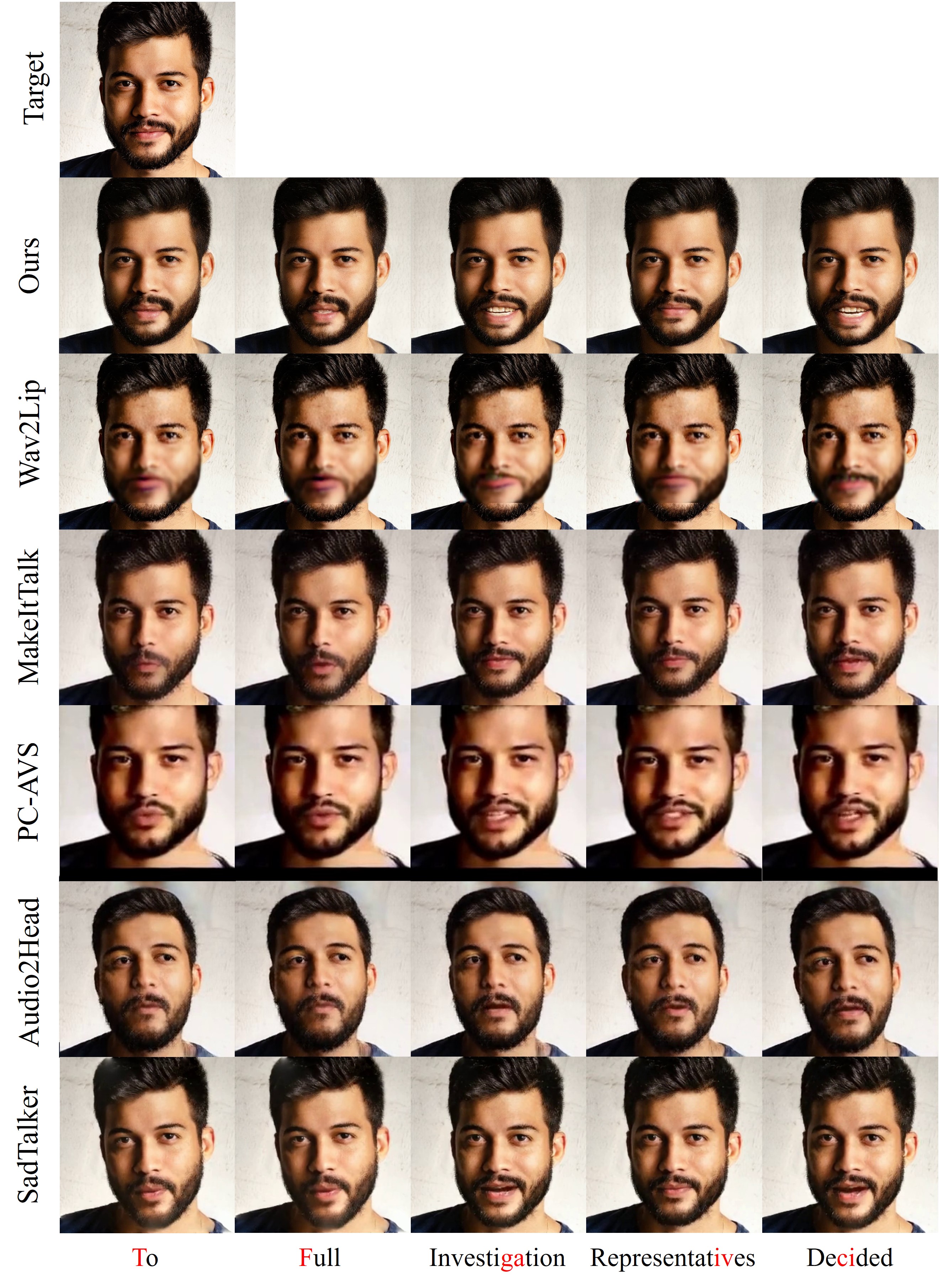}
   \vspace{-3mm}
   \caption{Visual comparison with outputs of baselines.}
   \vspace{-3mm}
   \label{fig:Compar_rot}
\end{figure}

\begin{figure*}[t]
\vspace{-3mm}
  \centering
   \includegraphics[width=0.90\textwidth]{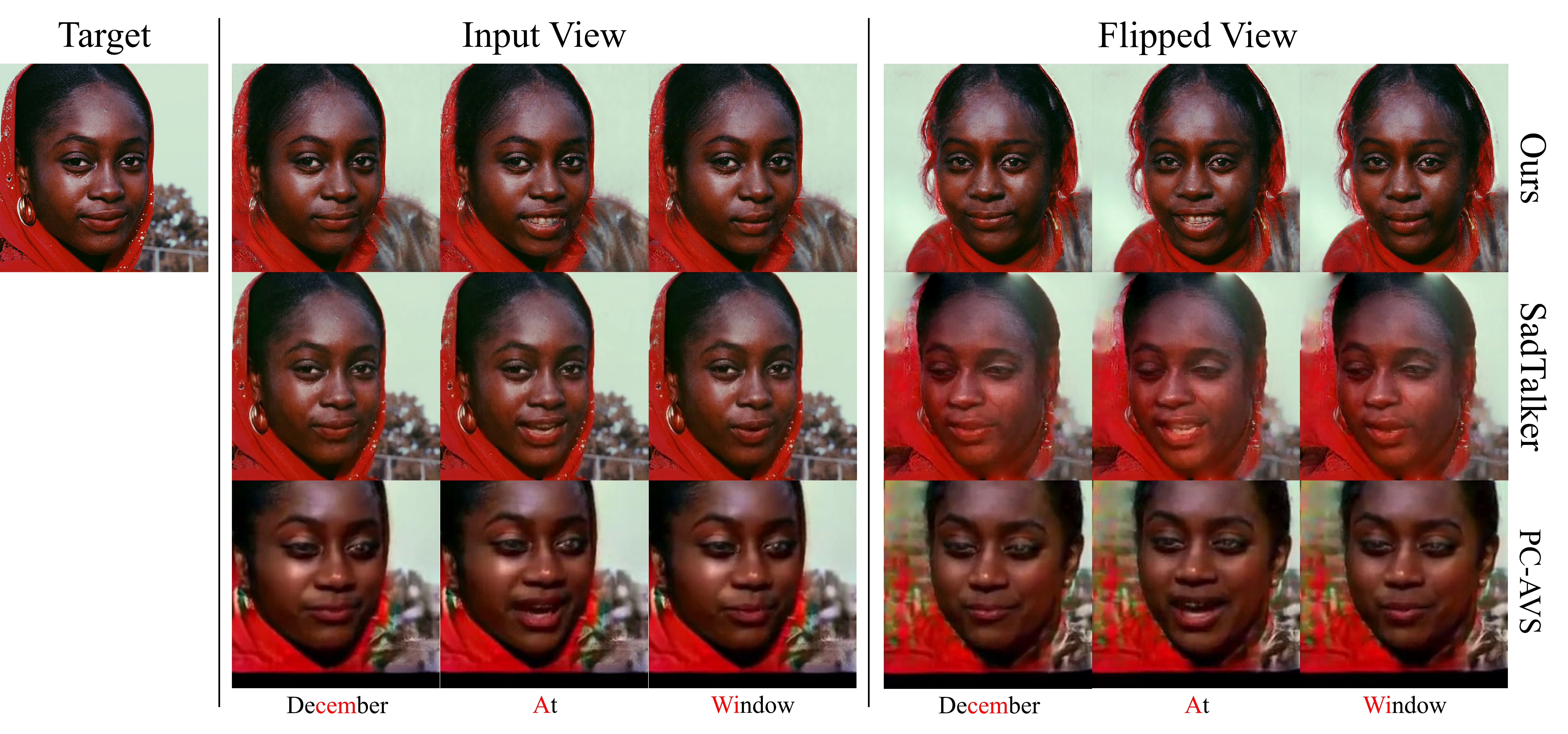}
   \vspace{-3mm}
   \caption{Visual comparison of the robustness to pose changes.}
   \vspace{-3mm}
   \label{fig:Compar}
\end{figure*}

\begin{table*}[t]
    \centering
    \resizebox{0.85\textwidth}{!}{
    \begin{tabular}{lc|ccccc|ccccc}
        \toprule
        \multirow{2}{*}{Method} & \multirow{2}{*}{Condition} & \multicolumn{5}{c|}{Unplash} & \multicolumn{5}{c}{HDTF} \\
        \cline{3-7} \cline{7-12}
          & & FID$\downarrow$ & CSIM$\uparrow$ & CPBD$\uparrow$ & LSE-D$\downarrow$ & LSE-C$\uparrow$ & FID$\downarrow$ & CSIM$\uparrow$ & CPBD$\uparrow$ & LSE-D$\downarrow$ & LSE-C$\uparrow$
         \\
        \hline
        \rowcolor{gray!20}
        PC-AVS & w/ input pose & 50.223 & 0.620  & 0.157  & 7.018 & 8.007 & 39.764  &0.699  & 0.148 & 6.501  & 8.738 \\
        \rowcolor{gray!20}
        PC-AVS & w/ h-flipped pose & 58.850 & 0.561  & 0.175  & 6.792 & 8.390 & 44.981  & 0.665 & 0.157 & 6.461 & 8.918 \\
  
         & \textbf{$\Delta$}  & -8.627  &  \textbf{0.058}  & -0.018  & 0.225  & -0.383 & -5.218 &  \textbf{0.034}  & -0.009  & 0.039  & -0.179 \\
        \hline

        \rowcolor{gray!20}
        SadTalker & w/ input pose & 54.953 & 0.914 & 0.166 & 7.989 & 6.913 & 14.702 & 0.959 & 0.147 & 7.618 & 7.463 \\
        \rowcolor{gray!20}
        SadTalker & w/ h-flipped pose & 70.640 & 0.793 & 0.195 & 8.073 & 6.788 & 23.710 & 0.885 & 0.171 & 7.764 & 7.260 \\
        
         & \textbf{$\Delta$}  & -15.687 & 0.122 & -0.030 & -0.084 & 0.124 & -9.009 & 0.074 & -0.024 & -0.146 & 0.203\\
        \hline
        
        \rowcolor{gray!20}
        Ours & w/ input pose & 50.463 & 0.747 & 0.581 & 8.869 & 5.941 & 36.994 & 0.826 & 0.225 & 9.194 & 5.729 \\
        \rowcolor{gray!20}
        Ours & w/ h-flipped pose & 54.403 & 0.671 & 0.574 & 8.866 & 5.937 & 44.363 & 0.750 & 0.229 & 9.160 & 5.815 \\
        
         & \textbf{$\Delta$}  & \textbf{-3.939} & 0.076 & \textbf{0.006} & \textbf{0.002} & \textbf{0.004} & \textbf{-7.369} & 0.077 & \textbf{-0.004} & \textbf{0.034} & \textbf{-0.086} \\
        \toprule
    \end{tabular}}
    \vspace{-3mm}
    \caption{Internal differences under pose conditions. We evaluate the changes in performance metrics within the same model when the conditions are altered. This includes measuring the differences $\Delta$ in outcomes when generating results using the original pose of the input image versus a horizontally flipped pose. The robustness of a model to pose changes is indicated by minimal variations in metrics between these two conditions (Bold: highest).} 
    \vspace{-3mm}

    \label{tab:comparison}
\end{table*}

\subsection{Evaluation Metrics}
\vspace{-1mm}
We validated the image quality and audio-visual multimodality of our experimental results using a set of evaluation metrics employed in prior studies. To assess the image quality of frames, we utilized the Fréchet Inception Distance (FID) \cite{heusel2017gans} that compares the distribution of our generated images against the original frames, the Cumulative Probability Blur Detection (CPBD) \cite{narvekar2011no} to gauge the sharpness of the images, and the Cosine Similarity Identity Metric (CSIM) \cite{deng2019arcface} to measure the preservation of identity between the source images and the resultant videos. Furthermore, to evaluate the lip-sync quality corresponding to audio inputs, we employed a Distance score (LSE-D) and a Confidence score (LSE-C), both derived from Wav2Lip \cite{prajwal2020lip}. We additionally introduce a novel evaluation metric referred to as Internal Differences (ID), which measures the difference in outcomes from the left-right mirrored poses of the input image. This metric indicates how consistently the model performs under different pose conditions.

\subsection{Comparisons with Baseline Methods}\label{sec:quan}
\vspace{-1mm}
We conducted two types of evaluations: a general performance and a focused analysis of pose changes. Initially, we compared our method with the recent state-of-the-art baselines \cite{prajwal2020lip,zhou2020makelttalk,wang2021audio2head, zhou2021pose, zhang2023sadtalker} on the Unplash and HDTF datasets for the general performance. Visual results are presented in Figure \ref{fig:Compar_rot}. Wav2Lip \cite{prajwal2020lip} produced results with the blurry mouth region and some artifacts in the beard region. Audio2Head \cite{wang2021audio2head} forced the target pose to be front facing, resulting in the change of  the subject's identity. This is also verified by the lowest scores achieved by Audio2Head in ID-preservation metrics as reported in \cref{sec:User}. 
An inaccurate lip sync was observed in MakeItTalk \cite{zhou2020makelttalk}, while PC-AVS \cite{zhou2021pose} suffers from degradation in the quality of generated results. Although SadTalker \cite{zhang2023sadtalker} achieved more natural-looking results, it showed a vulnerability to pose changes, as clearly illustrated in Figure \ref{fig:Compar}.

Despite demonstrating superior perceptual performance, as shown by the user study results reported in Table \ref{tab:user_study}, the quantitative results reported in Table \ref{tab:comparison_Unpalsh} indicate mid-tier numerical performance of our method in the evaluation metrics. One possible reason could be errors from the inversion process required for leveraging the backbone, which seems to particularly challenge the reconstruction, especially in background component, as illustrated in Figure \ref{fig:Compar}. Another possible reason, as observed in Tables \ref{tab:comparison_Unpalsh} and \ref{tab:user_study}, is the lack of consistency between the rankings of quantitative evaluation metrics and actual perceptual quality. This inconsistency has also been discussed in previous studies \cite{zhang2023sadtalker, wang2022one}, and further experiments detailed in Sec. \textcolor{red}{F} in the supplementary material demonstrate the inadequacy of existing evaluation metrics to accurately capture the quality of generated results.

Thus, our focus shifts from external comparisons between models to internal differences based on varying conditions. We examined the effects of using input views versus its horizontally flipped views on performance, as detailed in Table \ref{tab:comparison}. The use of these flipped pose inputs induces rotated outcomes from the input images, allowing us to measure the effectiveness of rotation generation. The comparison with pose-controllable baselines \cite{zhang2023sadtalker,zhou2021pose} reveals that our approach achieves minimal variance in most performance metrics under various pose conditions. This suggests a superior robustness to pose changes in our approach.

\subsection{User Study}\label{sec:User}
\vspace{-1mm}
In order to evaluate the general visual quality of output videos and assess the robustness of the pose variation, we conducted two user studies. Both studies involved 15 samples from the outputs. These samples of both studies were randomly shuffled and presented to 20 participants, comprising individuals in their 20s and 30s. They were asked to select the best sample based on the evaluation metrics. The result of the first user study is shown in Table \ref{tab:user_study}. We compared our method with baselines \cite{prajwal2020lip,zhou2020makelttalk,wang2021audio2head,zhang2023sadtalker}, which utilized the same processing as ours, using the Unplash and HDTF datasets. Our approach demonstrated superior performance across all metrics, notably in mouth quality, identity preservation, and video sharpness. This excellence is attributed to the effective utilization of ray deformation and LipaintNet. Ray deformation promotes natural movements without substituting features, thereby enabling motions free from identity shifts and degradation in image quality. Simultaneously, LipaintNet is responsible for generating inner-mouth details that faithfully reflect the high fidelity of the generative model, as demonstrated in Figure \ref{fig:Compar_rot}.

Next, we conducted a qualitative evaluation on off-frontal image data, a setup necessitating substantial rotation for horizontally flipped poses. Consequently, participants were tasked with assessing the robustness of generated rotations in response to these inputs. This assessment compared our method with SadTalker \cite{zhang2023sadtalker}. As shown in Table \ref{tab:user_study_rot}, our method maintained view consistency and showed strengths across all metrics while SadTalker struggled with identity preservation and exhibited transparency in facial regions when pose variations are significant. Specifically, SadTalker tended to produce results with blurry regions and identity shifts in the flipped pose results. In contrast, our method demonstrated robustness in handling these variations, maintaining both clarity and identity fidelity.

\begin{table}[t]
    \centering
    \resizebox{0.42\textwidth}{!}{
    \begin{tabular}{lcccc}
        \toprule
        Method & Lip Sync. & Mouth Quality & ID-Preserv. & Video Sharpness  \\
        \hline
        Wav2Lip \cite{prajwal2020lip}  & 29.7\% & 7.4\% & 7.0\% & 6.7\%  \\
        MakeItTalk \cite{zhou2020makelttalk}  & 3.0\% & 3.0\% & 3.3\% & 2.7\% \\
        Audio2Head \cite{wang2021audio2head} & 8.3\% & 8.4\% & 4.7\% & 3.0\%  \\
        SadTalker \cite{zhang2023sadtalker} & 23.7\% & 30.4\% & 27.0\% & 17.7\% \\
        Ours & \textbf{35.3\%} & \textbf{50.8\%} & \textbf{58.0\%} & \textbf{70.0\%} \\
        \toprule
    \end{tabular}
    }
    \vspace{-3mm}
    \caption{User study for general quality (Bold: highest).}
    \vspace{-3mm}
    \label{tab:user_study}
\end{table}

\begin{table}[t]
    \centering
    \resizebox{0.42\textwidth}{!}{
    \begin{tabular}{lcccc}
        \toprule
         Method & Lip Sync. & Mouth Quality & ID-Preserv. & Video Sharpness \\
        \hline
        SadTalker \cite{zhang2023sadtalker} & 36.5\% & 28.0\% & 20.2\% & 9.0\%   \\
        Ours & \textbf{63.5\%} & \textbf{72.0\%} & \textbf{79.8\%} & \textbf{91.0\%}  \\
        \toprule
    \end{tabular}}
    \vspace{-3mm}
    \caption{User study for rotation robustness (Bold: highest).}
    \vspace{-3mm}
    \label{tab:user_study_rot}
\end{table}

\subsection{Ablation Study}
\begin{figure}[t]
  \centering
  \vspace{-2mm}
   \includegraphics[width=0.41\textwidth]{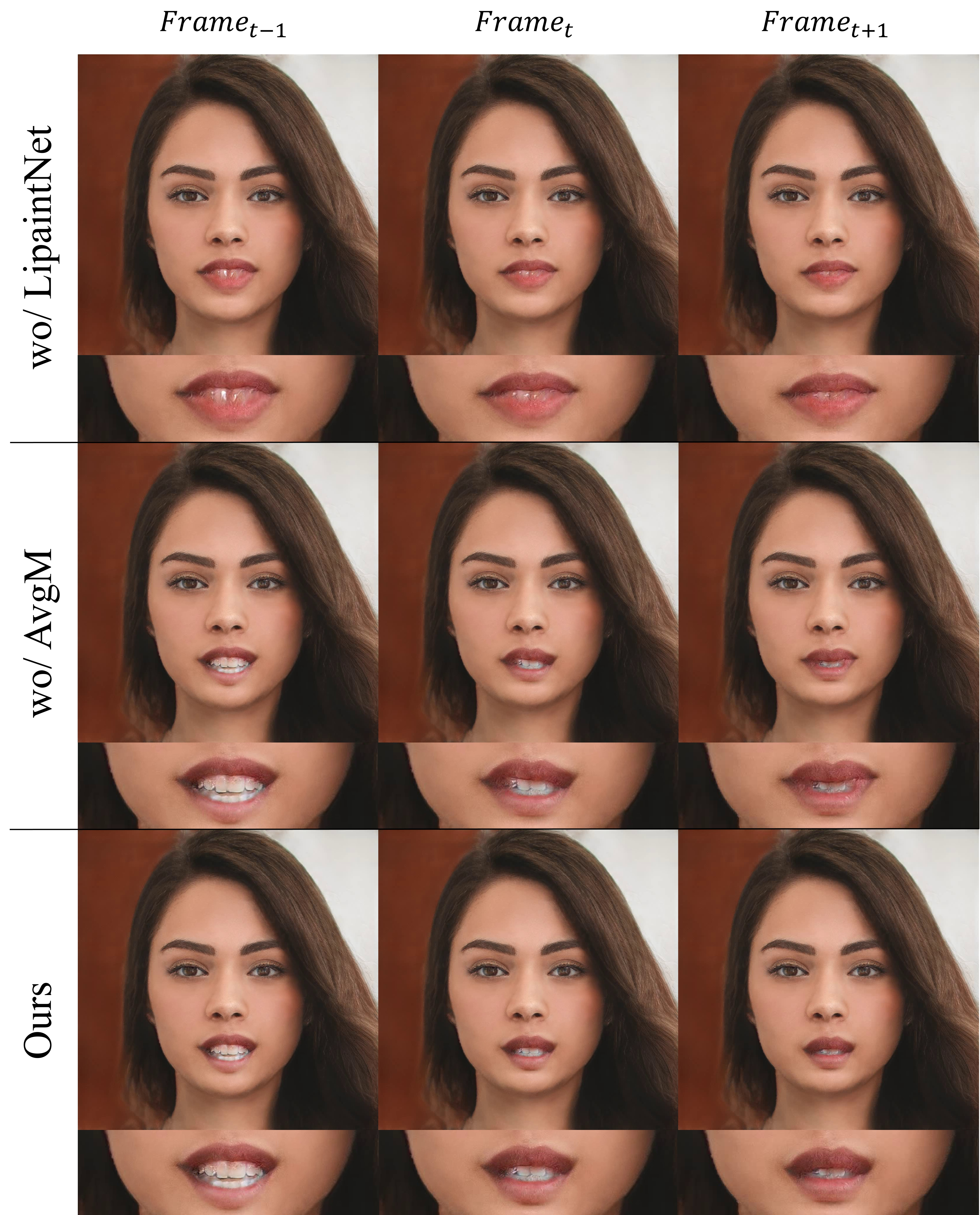}
   \vspace{-2mm}
   \caption{Ablation study for proposed LipaintNet and Average Mask methods.}
   \label{fig:abl}
\end{figure}

\vspace{-1mm}
To discern the contribution of each components of the proposed method to the overall performance, we conducted an ablation study. The study investigated four key elements: LipaintNet, which provides detailed inner-mouth information; the coefficients enhanced with motion dynamics through $\alpha_m$; expression scaling controlled by $\lambda_{exp}$; and Average Mask, responsible for calculating smooth transitions between frames. We evaluated performance of the model with and without these components. The results of this study is presented in Table \ref{tab:ablation_table}. It is evident that the absence of each component results in a decline in performance across almost all metrics, underscoring their positive influence on the quality of the final output. For instance, the first row of Figure \ref{fig:abl} shows that the mouth is inadequately filled. Without using the Average Mask resulted in lacking continuity and smooth in the final output, as shown in the second row. The last row shows that the mouth motion can be naturally generated with  smooth transitions across frames when all the components are integrated.

%% file: sec/6_conclusion.tex
\section{Conclusion}
\vspace{-1mm}
In this paper, we introduced NeRFFaceSpeech, a novel approach to generating 3D-aware audio-driven talking head animations from a single image by leveraging generative priors that construct and manipulate 3D features. Our pipeline bridges the gap between facial parametric models and neural rendering, intuitively manipulating the feature space through ray deformation. We also proposed LipaintNet, a self-supervised learning framework that leverages the capacity of the generative model for synthesizing the hidden inner-mouth area, complementing the deformation field to produce feasible results. Through extensive experimentation and user studies, we demonstrated that our model is robust to changes in pose and can generate superior internal mouth information, resulting in enhanced outcomes compared to previous methods.\\ \\
\noindent\textbf{Acknowledgement.} This work was supported by  Culture, Sports and Tourism R\&D Program through the KOCCA grant funded by the Ministry of Culture, Sports and Tourism (No. RS-2023-00228331)

\begin{table}[t]
    \centering
    \resizebox{0.42\textwidth}{!}{
    \begin{tabular}{lccccc}
        \toprule
        \multirow{2}{*}{Method} & \multicolumn{2}{c}{Unplash} & \multicolumn{2}{c}{HDTF} \\
        \cline{2-5} 
         & LSE-D$\downarrow$ & LSE-C$\uparrow$ & LSE-D$\downarrow$ & LSE-C$\uparrow$ \\
        \hline
        w/o LipaintNet  & 9.217 & 5.363 & 9.467 & 5.091  \\
        w/o $\alpha_m$  & 9.170 & 5.571 & 9.415 & 5.456  \\
        w/o $\lambda_{exp}$ & 9.169 & 5.377 & 9.255 & 5.516  \\
        w/o Average Mask & 8.924 & 5.917 & 9.223 & \textbf{5.800}  \\
        \hline
        Ours & \textbf{8.869} & \textbf{5.941} & \textbf{9.194} & 5.729 \\
        \toprule
    \end{tabular}
    }
    \vspace{-3mm}
    \caption{Quantitative results of ablation study (Bold: highest).}
    \vspace{-3mm}
    \label{tab:ablation_table}
\end{table}
